\documentclass[pmlr,twocolumn,10pt]{jmlr}

\mlhtrack{proceedings}

\newif\iffinal
\finaltrue

\iffinal
    \ifmlhneedspmlr
      \jmlrvolume{346} 
      \jmlryear{2026}
    \fi
    \ifmlhfindings \jmlrproceedings{}{ML4H 2026 - Findings Track}\fi
    \ifmlhdemo     \jmlrproceedings{}{ML4H 2026 - Demo Track}\fi
    \jmlrworkshop{Machine Learning for Health (ML4H) 2026}
\else
    \jmlrproceedings{}{Submitted to ML4H 2026: \mlhtrackname}
    \jmlrworkshop{Machine Learning for Health (ML4H) 2026}
\fi

\usepackage[utf8]{inputenc}
\usepackage[T1]{fontenc}
\usepackage{booktabs}
\usepackage{enumitem}
\usepackage{array}
\usepackage{microtype}
\usepackage{xcolor}
\usepackage{tikz}
\usetikzlibrary{arrows.meta,calc,positioning}
\usepackage{pgfplots}
\pgfplotsset{compat=1.18}

\usepackage[switch]{lineno}
\usepackage[capitalize]{cleveref}
\usepackage{listings}
\crefname{section}{Sec.}{Secs.}
\Crefname{table}{Table}{Tables}
\crefname{figure}{Fig.}{Figs.}

\title[Discharge Summary Generation Is Not Just Summarization]{%
Discharge Summary Generation Is Not Just Summarization: 
A Reality Check for Evaluation}

\author[Dua, Wolff, Cho, and Oermann]{
\Name{Radhika Dua}\,$^{*,1,2}$ \hfill \Email{radhikadua@nyu.edu}\\
\Name{Frederick Wolff}\,$^{*,2,3}$ \hfill \Email{wfrederick@ethz.ch}\\
\Name{Kyunghyun Cho}\,$^{1}$ \hfill \Email{kyunghyun.cho@nyu.edu}\\
\Name{Eric Karl Oermann}\,$^{1,2}$ \hfill \Email{eric.oermann@nyulangone.org}\\[4pt]
\addr{$^{1}$ Courant Institute School of Mathematics, Computing, and Data Science, New York University\\
$^{2}$ Department of Neurosurgery, NYU Langone Health\\
$^{3}$ Department of Computer Science, ETH Zurich\\
$^{*}$ These authors contributed equally and share first authorship.}
}

\iffinal\else
    \linenumbers
\fi

\begin{document}
\maketitle

\ifmlhdemo\else
\begin{figure*}[!t]
 \vspace{-1mm}
    \centering
    \resizebox{0.95\textwidth}{!}{%
        \input{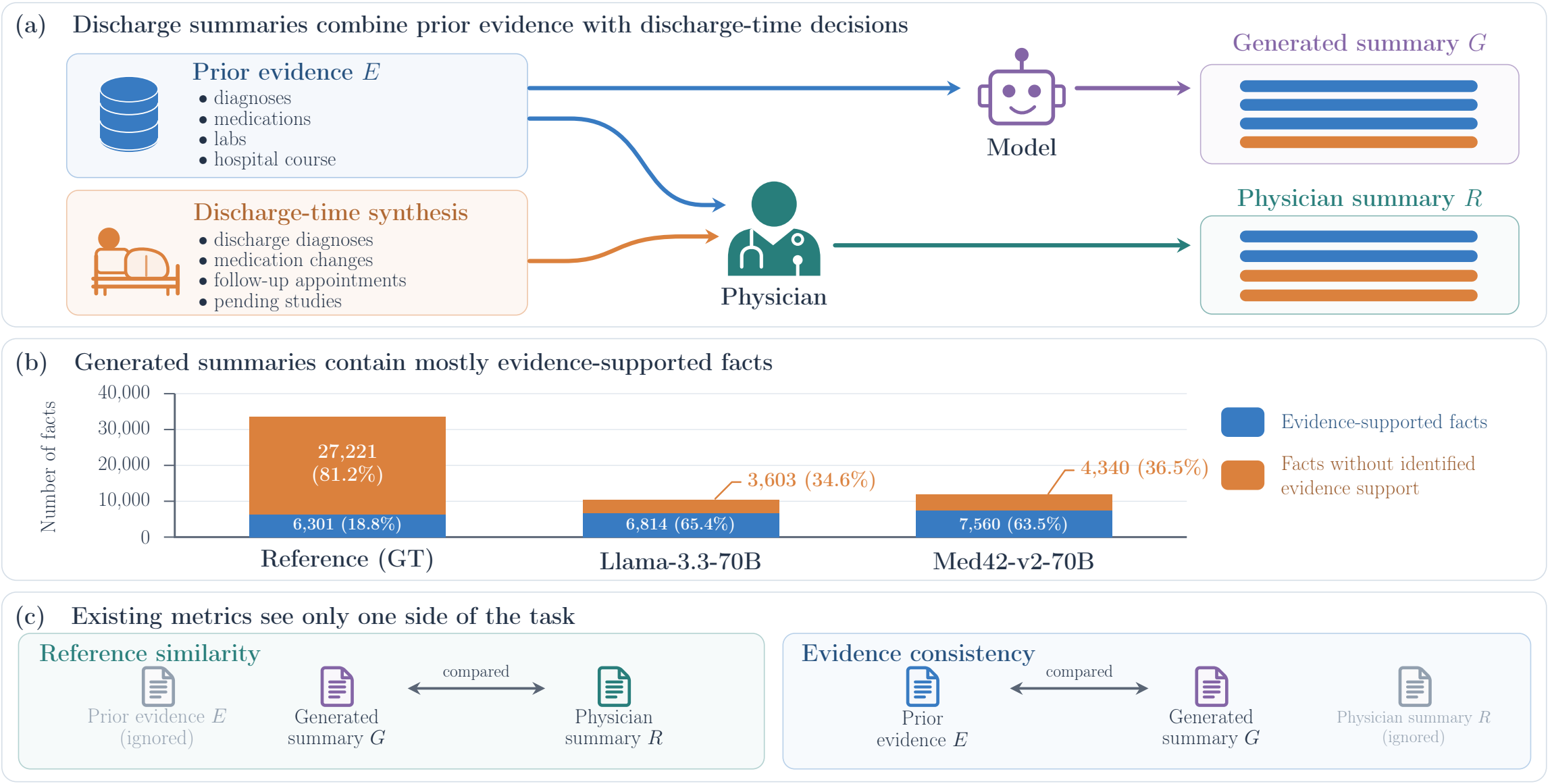}
    }
    \vspace{-3mm}
    \caption{\textbf{Why reference-only evaluation is misleading for discharge-summary generation.}
\textbf{(a)} Physician-written summaries $R$ combine prior evidence $E$
with discharge-time and inference-heavy content, whereas generated
summaries $G$ typically use $E$ alone, making them asymmetric
evaluation targets.
\textbf{(b)} References and model outputs contain comparable numbers
of evidence-supported facts, but references contain far more facts
not explicitly supported by prior evidence. A summarizer perfectly reproducing all evidence-supported content would still recover only 18.8\%
 of reference facts.
\textbf{(c)} Existing metrics see only one side of the task:
reference-similarity metrics compare $G$ with $R$ while ignoring $E$,
whereas evidence-consistency metrics compare $G$ with $E$ while
ignoring $R$. Standard evaluation can therefore conflate missed
evidence-grounded summarization with missed discharge-time inference
and future-care synthesis.}
 \vspace{-3mm}
    \label{fig:motivation}
\end{figure*}    

\begin{figure*}[!t]
    \vspace{-2mm}
    \centering
    \includegraphics[width=\textwidth]{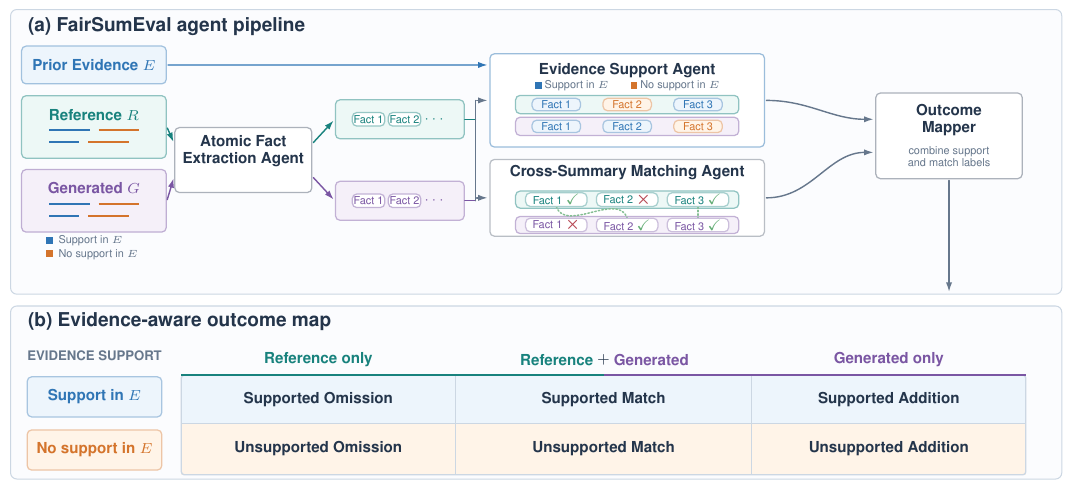}
    \vspace{-7mm}
    \caption{\textbf{Overview of FairSumEval.}
    \textbf{(a)} Given prior evidence $E$, physician-written reference $R$,
    and generated summary $G$, the Atomic Fact Extraction Agent extracts
    facts from both summaries. The Evidence Support and Cross-Summary
    Matching Agents assess support in $E$ and coverage by the opposite
    summary, respectively.
    \textbf{(b)} The Outcome Mapper combines these decisions with fact
    origin to distinguish omissions, matches, and additions within
    the support in $E$ and no support in $E$ strata. Match cells retain
    directional coverage counts without assuming one-to-one
    correspondences. }
     \vspace{-3mm}
    \label{fig:methodv4}
\end{figure*}

\begin{abstract}
Discharge summaries are hybrid clinical documents: they summarize what happened during a hospitalization and communicate discharge-time decisions about what should happen next. This makes their evaluation fundamentally different from standard summarization. Physician-written references contain both facts supported by prior clinical notes and reference-only content, potentially reflecting clinical synthesis, handoffs, or discharge planning. Evaluating a generated summary only against the reference can therefore 
conflate failure to recover available information with failure to reproduce content not identifiable in the model input.
We introduce \textsc{FairSumEval}, an evidence-aware, fact-level framework that asks two questions about every fact in the physician reference and generated summary: is the fact supported by the evidence available to the model, and is it captured by the other summary?
The answers produce a complete outcome map of matches, omissions, and additions.
We summarize this map using two complementary metrics: Soft IoU-S measures
reference--generation overlap among facts supported by $E$, while Soft IoU-NS
measures overlap among facts without identified support in $E$. On 200 MIMIC-III admissions, only 18.8\% of physician reference facts have identified support in the prior evidence, compared with 65.4\% and 63.5\% of facts generated by a general-purpose and a medically specialized model. Across both models, overlap is nearly an order of magnitude higher for supported content (0.255 and 0.236) than for content without identified support (0.030 and 0.026). However, the two scores are only weakly correlated across admissions ($\rho=0.22$ and $0.17$), showing that performance on one does not predict performance on the other. Conventional metrics also fail to recover this distinction: lexical metrics align more strongly with Soft IoU-NS, and no evaluated metric consistently captures either dimension. By separating these outcomes, \textsc{FairSumEval} distinguishes evidence-grounded summarization from agreement on inference-heavy and future-care content, while revealing what models omit and add in each category. 
\end{abstract}

\begin{keywords}
Discharge summaries, clinical text generation, evidence-aware evaluation,
large language models
\end{keywords}
\fi

\ifmlhneedsstatements
\paragraph*{Data and Code Availability}
This paper uses the MIMIC-III dataset,
available through PhysioNet to credentialed researchers.
Code is available at
\url{https://github.com/nyuolab/FairSumEval}

\paragraph*{Institutional Review Board (IRB)}
This research does not require IRB approval.
\fi

\vspace{-2mm}
\section{Introduction}

Discharge summaries are not ordinary summaries. They retrospectively
describe what happened during a hospitalization and prospectively
communicate what should happen after discharge~\citep{Snow2009TransitionsOC}.
The first role requires distilling diagnoses, treatments, and hospital
events from the clinical record~\citep{Adams2021WhatsIA}. The second
can involve clinical inference and discharge-time decisions, including
medication changes and follow-up plans that may not be documented
in prior notes. As illustrated in Figure~\ref{fig:motivation}a,
physician-written summaries combine these sources of information,
whereas generation models may receive only the prior clinical
evidence. As language models are increasingly used to draft these
documents~\citep{Socrates2024YaleA,Aali2024ADA}, evaluation must account
for both the information available to the model and the content
expected in its output.

Conventional reference-based evaluation compares a generated summary
$G$ with a physician-written reference $R$ using lexical, semantic,
or clinical concept overlap~\citep{Socrates2024YaleA,Adams2023AMO,Lin2004ROUGEAP,Papineni2002BleuAM,Zhang2019BERTScoreET}.
However, interpreting this agreement as summarization performance
overlooks whether the reference content is supported by the model's
input $E$. Prior work has identified content without source support
in clinician-written hospital-course references~\citep{Adams2022LearningTR}.
A missing reference fact may therefore reflect either failure to
recover available information or failure to reproduce content without
identified support in the input. Conversely, a generated fact absent
from the reference may be supported by the evidence and potentially
useful. Figure~\ref{fig:motivation}c illustrates the resulting
evaluation gap: reference similarity compares $G$ with $R$ without
considering $E$, whereas evidence consistency compares $G$ with $E$
without considering $R$. Neither comparison alone establishes
whether the generation captures the supported content selected by
the physician. Without these distinctions, higher scores are
difficult to interpret as progress toward better summarization.

We introduce \textsc{FairSumEval}, an evidence-aware, fact-level
framework that evaluates the full triple $(E,R,G)$. For each atomic
fact extracted from the reference and generated summary, it asks
two questions: \emph{Is the fact supported by the available evidence?}
and \emph{Is it covered by the other summary?} Together, these
decisions produce a complete outcome map of matches, omissions,
and additions within two strata: support in $E$ and no identified
support in $E$. We summarize this map using two complementary
metrics: Soft IoU-S measures overlap among supported facts, and
Soft IoU-NS measures overlap among facts without identified support.
The scores are reported separately to distinguish these dimensions
of agreement. Absence of identified support does not imply clinical
incorrectness; it may reflect clinical synthesis or information
outside the supplied evidence.

We evaluate \textsc{FairSumEval} on 200 MIMIC-III admissions using
summaries generated by Llama-3.3-70B and Med42-v2-70B.
Figure~\ref{fig:motivation}b reveals a sharp difference in evidence
composition: fewer than one in five physician-reference facts have
identified support in the prior notes, compared with roughly two
in three generated facts. Both models agree with the reference
much more closely on supported content than on content without
identified support. However, admissions with stronger agreement
on supported content do not consistently show stronger agreement
on the remaining content, indicating that the two scores capture
different aspects of generation. Conventional metrics obscure this
distinction: ROUGE-L and BLEU-4 align more closely with agreement
on content without identified support, while BERTScore and
SummaC-ZS provide no consistent substitute across models.

These findings motivate a clearer evaluation target for
discharge-summary generation. Improving agreement on available
information and reproducing content beyond that information are
different objectives, potentially requiring different methods
or additional inputs. \textsc{FairSumEval} makes this distinction
explicit, providing a more interpretable basis for assessing
what a method improves and where further development is needed.
Our contributions are:
\begin{itemize}[nosep]
    \item We identify and quantify an asymmetry in evidence support
    between physician-written references and generated discharge
    summaries that complicates reference-based evaluation.
    \item We introduce \textsc{FairSumEval}, a complete evidence-aware
    outcome map summarized by Soft IoU-S and Soft IoU-NS.
    \item We show that these scores capture nonredundant dimensions
    of agreement that the evaluated conventional metrics do not
    consistently recover.
\end{itemize}
\vspace{-2mm}
\section{Method: FairSumEval}

\textsc{FairSumEval} evaluates reference--generation agreement
conditional on the evidence available to the model.
As shown in Figure~\ref{fig:methodv4}, it extracts clinical facts,
assesses evidence support and cross-summary coverage, and maps
these decisions to omissions, matches, and additions within two
evidence strata. The complete outcome map is the primary output;
IoU-S and IoU-NS summarize overlap within each stratum. RAGChecker also jointly considers references, generations, and
evidence~\citep{Ru2024RAGCheckerAF}; our focus is on clinician-written
references only partially supported by available prior notes.

\vspace{-1mm}
\subsection{Problem Setup and Evidence-Aware Decomposition}

For each admission, let $E$ denote the clinical notes available
to the model, $R$ the physician-written discharge summary, and
$G$ the generated summary. We omit the admission index for clarity.
Rather than evaluating $M(G,R)$ alone, we evaluate $M(E,R,G)$
through two binary decisions for each extracted fact: whether
$E$ supports it and whether the opposite summary covers it.
Combining these decisions with fact origin yields the outcome map.

Throughout, ``no support in $E$'' means that no support is
identified in the supplied evidence. It does not establish clinical
incorrectness or distinguish clinical synthesis from information
absent from $E$.

\vspace{-1mm}
\subsection{Atomic Fact Extraction}

An extraction agent $\Phi$ independently decomposes the summaries
into fact sets $F_R=\Phi(R)$ and $F_G=\Phi(G)$, following precedents
in general and clinical factuality
evaluation~\citep{Min2023FActScoreFA,Hu2024KnowledgeCentricHD,Munnangi2024FactEHRAD}.
Each fact expresses a self-contained clinical proposition.
The prompt splits multi-fact sentences while preserving negation,
uncertainty, diagnoses, medications, procedures, values, dates,
and follow-up plans. The complete extraction prompt and structured input and output formats are
provided in Appendix~\ref{app:fact-extraction-prompt}.

\vspace{-1mm}
\subsection{Evidence-Support Assessment}

For each fact $f\in F_R\cup F_G$, a binary verifier $v(f,T)$
returns $1$ when text $T$ is judged to fully support its factual
content, and $0$ otherwise. Evidence support is $s(f,E)=v(f,E)$;
when $E$ is chunked, support is assigned if any chunk fully
supports the fact. This assessment requires direct support:
clinical plausibility, external knowledge, or unstated
discharge-time judgments alone do not qualify. The complete evidence support prompt and structured interface are provided in
Appendix~\ref{app:evidence-support-prompt}.  

\vspace{-1mm}
\subsection{Cross-Summary Matching}

The same verifier assesses coverage by the opposite summary,
independently of the evidence-support label:
\begin{equation}
\begin{aligned}
    m_R(r,G) &= v(r,G), && r\in F_R,\\
    m_G(g,R) &= v(g,R), && g\in F_G.
\end{aligned}
\end{equation}
Matching uses the full opposite summary and requires support
for the fact's content. It is directional and imposes no
one-to-one correspondence; the numbers of covered reference
and generated facts can therefore differ. The complete directional matching prompt and structured interface are provided
in Appendix~\ref{app:cross-summary-prompt}.

\subsection{Outcome Assignment and Metric Aggregation}

We partition each fact set by evidence support. For
$X\in\{R,G\}$, define
\begin{equation}
\begin{aligned}
    X^+ &= \{f\in F_X:s(f,E)=1\},\\
    X^- &= \{f\in F_X:s(f,E)=0\}.
\end{aligned}
\end{equation}

\paragraph{Evidence-aware outcome map.}
For each stratum $\sigma\in\{+,-\}$, the directional match counts are
\begin{equation}
\begin{aligned}
    M_{R\rightarrow G}^{\sigma}
    &= \sum_{r\in R^\sigma}m_R(r,G),\\
    M_{G\rightarrow R}^{\sigma}
    &= \sum_{g\in G^\sigma}m_G(g,R).
\end{aligned}
\end{equation}
Unmatched reference facts are omissions, and unmatched generated
facts are additions:
\begin{equation}
\begin{aligned}
    O^\sigma &= |R^\sigma|-M_{R\rightarrow G}^\sigma,\\
    A^\sigma &= |G^\sigma|-M_{G\rightarrow R}^\sigma.
\end{aligned}
\end{equation}
These counts define the six cells in Figure~\ref{fig:methodv4},
with each match cell retaining both directions. Evidence status
belongs to the assessed fact, while coverage is checked against
the full opposite summary. The map thus reports directional
coverage rather than a deduplicated union of facts.

\paragraph{IoU-S and IoU-NS.}
We average the directional match counts and define stratum-specific
soft overlap:
\begin{equation}
\begin{aligned}
    M_{\mathrm{soft}}^\sigma
    &= \frac{M_{R\rightarrow G}^\sigma
             +M_{G\rightarrow R}^\sigma}{2},\\
    \mathrm{IoU}^\sigma
    &= \frac{M_{\mathrm{soft}}^\sigma}
    {|R^\sigma|+|G^\sigma|-M_{\mathrm{soft}}^\sigma}.
\end{aligned}
\end{equation}
IoU-S is $\mathrm{IoU}^{+}$ (support in $E$), and IoU-NS is
$\mathrm{IoU}^{-}$ (no support in $E$). These are soft IoU-style
scores, referred to as Soft IoU-S and Soft IoU-NS in the results:
they summarize directional coverage without constructing unique
matched fact pairs.

Scores are computed per admission and generated summary.
A zero denominator makes that score undefined; it is excluded
only from analyses requiring it. Cohort means average defined
admission-level scores rather than pooled counts, and correlations
retain admissions where both compared measures are defined.

\vspace{-1mm}
\subsection{Interpretation of the Evidence-Aware Outcomes}

The supported stratum describes recovery and omission of available
reference content, together with supported additions beyond the
reference. The no-support stratum describes agreement and differences
over content without identified support in $E$; a match does not
itself validate the underlying inference. Supported additions may
be useful, while omissions without identified support need not
indicate failure to summarize available evidence. Higher overlap
therefore indicates closer cross-summary agreement, not necessarily
greater clinical correctness. We retain the complete outcome map
for interpretation and report IoU-S and IoU-NS separately rather
than combining them into a universal quality score.
\vspace{-2mm}
\section{Experiments and Results}
\label{sec:experiments}

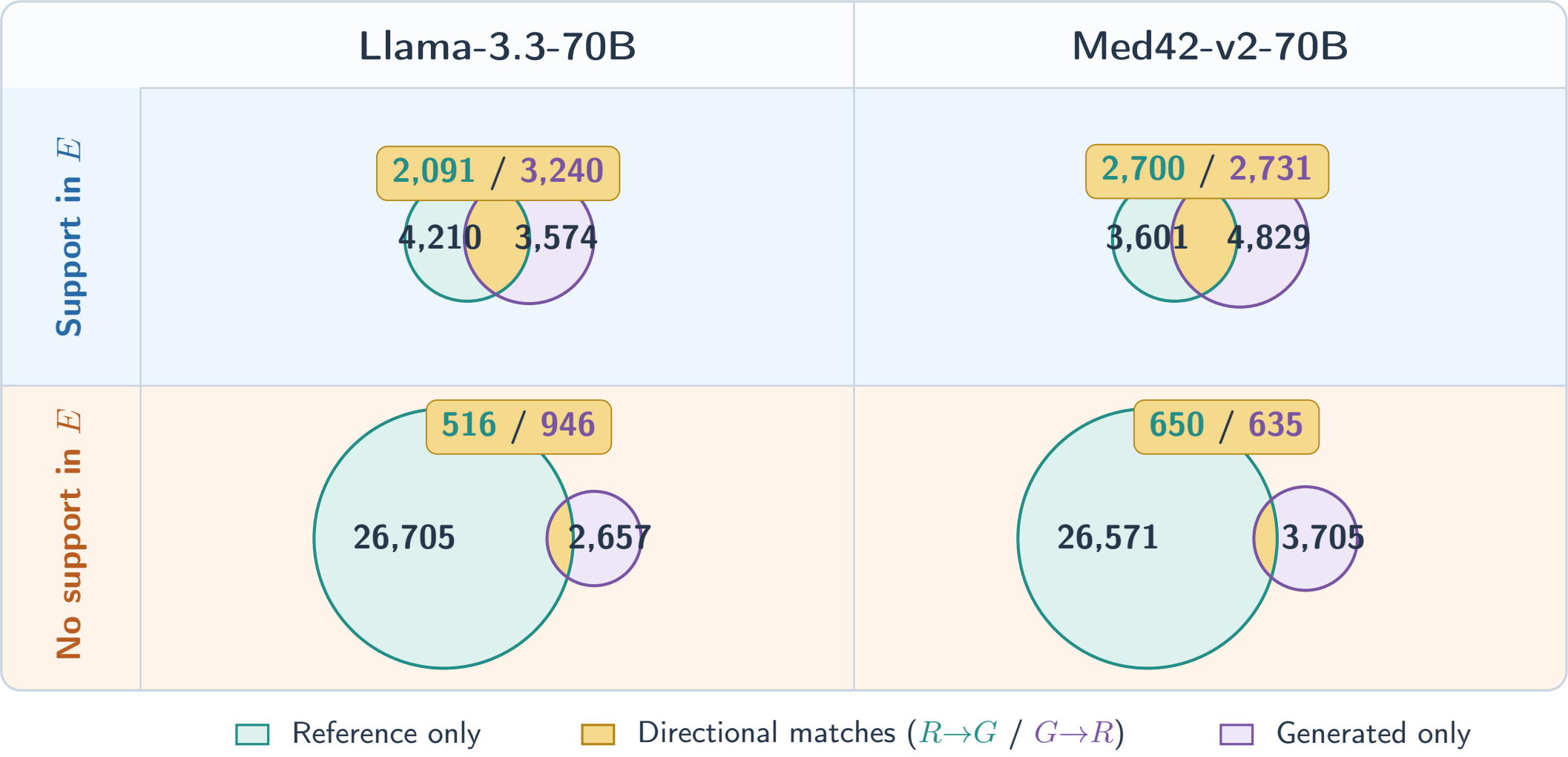
\begin{figure*}[!t]
     \vspace{-2mm}
    \centering
    \resizebox{0.95\textwidth}{!}{%

\definecolor{omInk}{HTML}{26364A}
\definecolor{omMuted}{HTML}{6E7A88}
\definecolor{omRule}{HTML}{CAD5E1}
\definecolor{omPanel}{HTML}{FBFCFE}
\definecolor{omSupportFill}{HTML}{EEF6FD}
\definecolor{omSupportLine}{HTML}{75AEE2}
\definecolor{omSupportInk}{HTML}{286AA6}
\definecolor{omNoSupportFill}{HTML}{FFF4EA}
\definecolor{omNoSupportLine}{HTML}{E9A36A}
\definecolor{omNoSupportInk}{HTML}{B85D22}
\definecolor{omRefFill}{HTML}{DDF1EF}
\definecolor{omRefLine}{HTML}{238D87}
\definecolor{omGenFill}{HTML}{EEE7F8}
\definecolor{omGenLine}{HTML}{7954A3}
\definecolor{omMatchFill}{HTML}{F5D98D}
\definecolor{omMatchLine}{HTML}{B48616}

\begin{tikzpicture}[
    x=1cm,
    y=1cm,
    every node/.style={font=\sffamily,text=omInk},
    model/.style={font=\sffamily\bfseries\large,align=center},
    rowtitle/.style={font=\sffamily\bfseries\normalsize,align=center},
    count/.style={font=\sffamily\bfseries\normalsize,align=center},
    matchcount/.style={rounded corners=1.1mm,draw=omMatchLine,
                      fill=omMatchFill,inner xsep=4.5pt,inner ysep=3pt,
                      font=\sffamily\bfseries\normalsize,align=center},
    legend/.style={font=\sffamily\small,anchor=west},
    divider/.style={draw=omRule,line width=.55pt}
]

\begin{scope}
  \clip[rounded corners=2mm] (.25,.18) rectangle (16.55,7.35);
  \fill[omPanel] (.25,.18) rectangle (16.55,7.35);
  \fill[omNoSupportFill] (.25,.18) rectangle (16.55,3.35);
  \fill[omSupportFill] (.25,3.35) rectangle (16.55,6.45);
\end{scope}
\draw[rounded corners=2mm,draw=omRule,line width=.7pt]
  (.25,.18) rectangle (16.55,7.35);

\draw[divider] (1.70,6.45) -- (16.55,6.45);
\draw[divider] (1.70,.18) -- (1.70,6.45);
\draw[divider] (9.13,.18) -- (9.13,7.35);
\node[model] at (5.42,6.90) {Llama-3.3-70B};
\node[model] at (12.84,6.90) {Med42-v2-70B};

\draw[divider] (.25,3.35) -- (16.55,3.35);
\node[rowtitle,text=omSupportInk,rotate=90] at (.98,4.90)
  {Support in $E$};
\node[rowtitle,text=omNoSupportInk,rotate=90] at (.98,1.80)
  {No support in $E$};

\begin{scope}
  \fill[omRefFill] (5.103,4.88) circle[radius=.650cm];
  \fill[omGenFill] (5.747,4.88) circle[radius=.675cm];
  \begin{scope}
    \clip (5.103,4.88) circle[radius=.650cm];
    \fill[omMatchFill] (5.747,4.88) circle[radius=.675cm];
  \end{scope}
  \draw[omRefLine,line width=.9pt] (5.103,4.88) circle[radius=.650cm];
  \draw[omGenLine,line width=.9pt] (5.747,4.88) circle[radius=.675cm];
\end{scope}
\node[count] at (4.82,4.88) {4,210};
\node[count] at (6.03,4.88) {3,574};
\node[matchcount] at (5.425,5.56)
  {\textcolor{omRefLine}{2,091} / \textcolor{omGenLine}{3,240}};

\begin{scope}
  \fill[omRefFill] (12.471,4.88) circle[radius=.650cm];
  \fill[omGenFill] (13.148,4.88) circle[radius=.711cm];
  \begin{scope}
    \clip (12.471,4.88) circle[radius=.650cm];
    \fill[omMatchFill] (13.148,4.88) circle[radius=.711cm];
  \end{scope}
  \draw[omRefLine,line width=.9pt] (12.471,4.88) circle[radius=.650cm];
  \draw[omGenLine,line width=.9pt] (13.148,4.88) circle[radius=.711cm];
\end{scope}
\node[count] at (12.19,4.88) {3,601};
\node[count] at (13.45,4.88) {4,829};
\node[matchcount] at (12.810,5.58)
  {\textcolor{omRefLine}{2,700} / \textcolor{omGenLine}{2,731}};

\begin{scope}
  \fill[omRefFill] (4.857,1.76) circle[radius=1.350cm];
  \fill[omGenFill] (6.424,1.76) circle[radius=.491cm];
  \begin{scope}
    \clip (4.857,1.76) circle[radius=1.350cm];
    \fill[omMatchFill] (6.424,1.76) circle[radius=.491cm];
  \end{scope}
  \draw[omRefLine,line width=.9pt] (4.857,1.76) circle[radius=1.350cm];
  \draw[omGenLine,line width=.9pt] (6.424,1.76) circle[radius=.491cm];
\end{scope}
\node[count] at (4.45,1.76) {26,705};
\node[count] at (6.59,1.76) {2,657};
\node[matchcount] at (5.640,2.92)
  {\textcolor{omRefLine}{516} / \textcolor{omGenLine}{946}};

\begin{scope}
  \fill[omRefFill] (12.187,1.76) circle[radius=1.350cm];
  \fill[omGenFill] (13.833,1.76) circle[radius=.539cm];
  \begin{scope}
    \clip (12.187,1.76) circle[radius=1.350cm];
    \fill[omMatchFill] (13.833,1.76) circle[radius=.539cm];
  \end{scope}
  \draw[omRefLine,line width=.9pt] (12.187,1.76) circle[radius=1.350cm];
  \draw[omGenLine,line width=.9pt] (13.833,1.76) circle[radius=.539cm];
\end{scope}
\node[count] at (11.78,1.76) {26,571};
\node[count] at (14.02,1.76) {3,705};
\node[matchcount] at (13.010,2.92)
  {\textcolor{omRefLine}{650} / \textcolor{omGenLine}{635}};

\fill[omRefFill] (2.70,-.38) rectangle ++(.33,.20);
\draw[omRefLine,line width=.7pt] (2.70,-.38) rectangle ++(.33,.20);
\node[legend] at (3.15,-.28) {Reference only};
\fill[omMatchFill] (6.30,-.38) rectangle ++(.33,.20);
\draw[omMatchLine,line width=.7pt] (6.30,-.38) rectangle ++(.33,.20);
\node[legend] at (6.75,-.28)
  {Directional matches
   (\textcolor{omRefLine}{$R{\rightarrow}G$} /
    \textcolor{omGenLine}{$G{\rightarrow}R$})};
\fill[omGenFill] (12.95,-.38) rectangle ++(.33,.20);
\draw[omGenLine,line width=.7pt] (12.95,-.38) rectangle ++(.33,.20);
\node[legend] at (13.40,-.28) {Generated only};

\end{tikzpicture}%
    }
     \vspace{-4mm}
    \caption{\textbf{Evidence-aware outcome maps for Llama-3.3-70B and Med42-v2-70B.}
    Rows separate facts with identified support in $E$ (blue) from those
    without (orange). Green, gold, and purple regions represent
    reference-only facts, matches, and generation-only facts, respectively.
    Circle areas scale with fact counts in each summary and evidence
    stratum; gold lens areas scale with the mean directional match count.
    Gold labels report directional match counts in the order $R{\rightarrow}G$ / $G{\rightarrow}R$, with colors corresponding to the reference and generation. These counts may differ because facts are
    extracted independently from $R$ and $G$. All counts are pooled
    across 200 admissions.}
    \label{fig:fairsum_outcome_map}
\end{figure*}

\begin{table*}[!t]
     \vspace{-3mm}
    \footnotesize
    \centering
    \caption{\textbf{Admission-level Soft IoU.}
Soft IoU-S and Soft IoU-NS measure overlap among facts with and
without identified evidence support, respectively. Columns report
means [95\% bootstrap CI] and medians $[Q_1,Q_3]$.
Both models show substantially closer agreement with the physician
reference on supported content than on content without identified support.}
\vspace{-2mm}
    \label{tab:admission-level-soft-iou-compact}
    \setlength{\tabcolsep}{5pt}
    \renewcommand{\arraystretch}{1.18}
    \begin{tabular}{lcccc}
        \toprule
        & \multicolumn{2}{c}{$\mathbf{Soft\ IoU\text{-}S}\;\boldsymbol{\uparrow}$}
            & \multicolumn{2}{c}{$\mathbf{Soft\ IoU\text{-}NS}\;\boldsymbol{\uparrow}$} \\
        \cmidrule(lr){2-3}\cmidrule(lr){4-5}
        \textbf{Generation model}
            & Mean
            & Median
            & Mean
            & Median \\
        \midrule
        Llama-3.3-70B
            & \textbf{0.255 {[0.240, 0.272]}}
            & \textbf{0.255 {[0.190, 0.307]}}
            & \textbf{0.030 {[0.026, 0.033]}}
            & \textbf{0.025 {[0.011, 0.042]}} \\
        Med42-v2-70B
            & 0.236 {[0.221, 0.252]}
            & 0.231 {[0.158, 0.297]}
            & 0.026 {[0.023, 0.029]}
            & 0.020 {[0.011, 0.034]} \\
        \bottomrule
    \end{tabular}
\end{table*}

\begin{figure*}[!t]
     \vspace{-2mm}
    \centering
    \includegraphics[
        width=0.7\textwidth,
        trim=0 0 0 0,
        clip
    ]{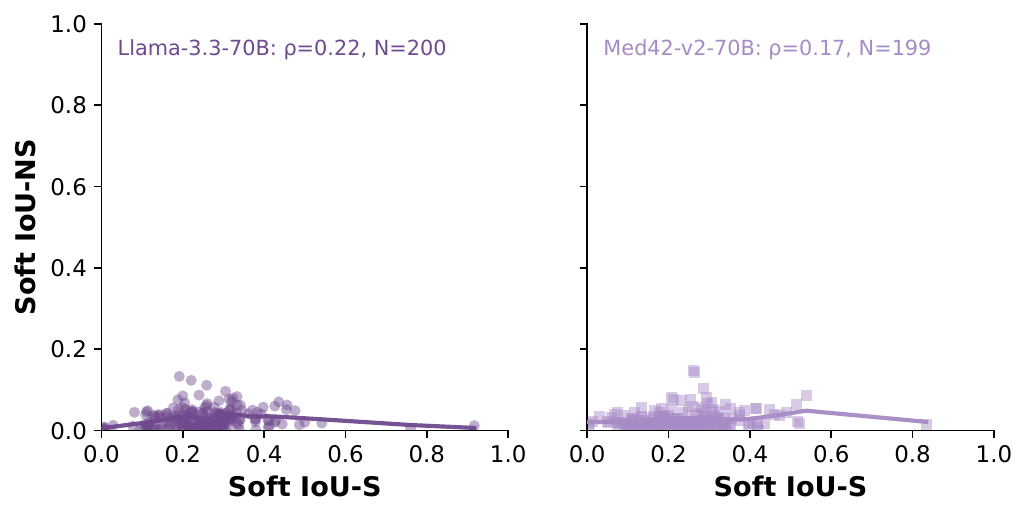}
    \vspace{-5mm}
\caption{\textbf{Evidence-aware evaluation reveals two non-redundant
dimensions of summary overlap.}
Each point represents one admission for Llama-3.3-70B (left) or
Med42-v2-70B (right). Soft IoU-S and Soft IoU-NS measure
reference--generation overlap among facts with and without identified
evidence support, respectively. Both models show substantially higher
supported overlap, but the two scores are only weakly associated
(Spearman $\rho=0.22$ and $0.17$, respectively).
Thus, stronger agreement
on supported content does not reliably predict greater agreement
on content without identified support, motivating separate reporting.
Trend lines are only visual aids.}
    \label{fig:soft-iou-relationship}
\end{figure*}

\begin{figure*}[!t]
 \vspace{-2mm}
    \centering
    \includegraphics[
        width=0.76\textwidth,
        trim=0 0 0 0,
        clip
    ]{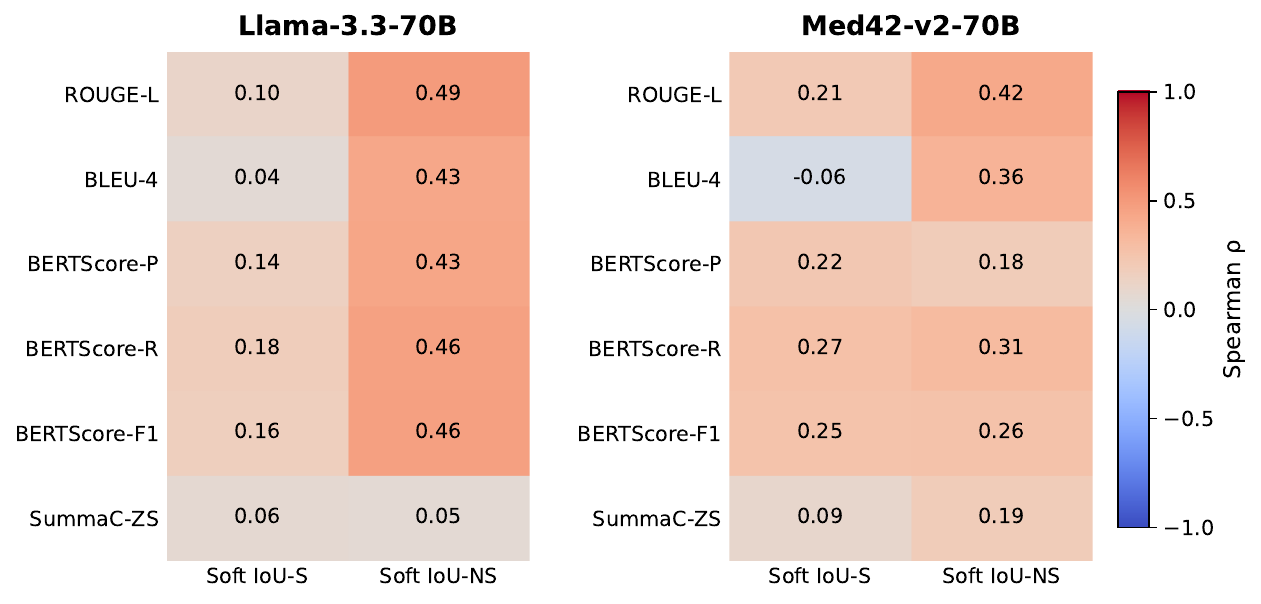}
 \vspace{-5mm}
\caption{\textbf{Conventional metrics do not consistently capture evidence-aware agreement.}
Each cell reports the admission-level Spearman correlation of a
metric with Soft IoU-S or Soft IoU-NS ($N=200$ for
Llama-3.3-70B; $N=199$--$200$ for Med42-v2-70B).
ROUGE-L and BLEU-4 correlate more strongly with Soft IoU-NS for
both models, indicating closer alignment with agreement on content
without identified evidence support. BERTScore shows this pattern
for Llama but no clear preference for either dimension for Med42;
SummaC-ZS is weakly associated with both. These metrics therefore
provide no consistent substitute for separating agreement by
evidence support.}
\vspace{-1mm}
\label{fig:conventional-metric-correlations}
\end{figure*}

We evaluate whether accounting for evidence availability changes what can be
learned from reference--generation agreement. We first compare the evidence
composition of physician-written summaries with that of summaries produced by
a general-purpose and a medically specialized model. We then use
\textsc{FairSumEval} to characterize their fact-level agreement, determine
whether Soft IoU-S and Soft IoU-NS capture distinct behaviors, and assess
whether these behaviors are reflected by conventional evaluation metrics.

\vspace{-1mm}
\subsection{Experimental Setup}
\label{sec:setup}

\paragraph{Data and generation models.}
We evaluate \textsc{FairSumEval} on 200 admissions from
MIMIC-III~\citep{Johnson2016MIMICIIIAF}. For each admission, the admission's clinical notes from nine MIMIC-III categories serve as evidence $E$ (Appendix~\ref{app:cohort-evidence}), and the
physician-written discharge summary serves as reference $R$. From the same evidence, we generate one summary using the general-purpose Llama-3.3-70B model\footnote{Official model card: \url{https://huggingface.co/meta-llama/Llama-3.3-70B-Instruct}.} and one using the medically specialized Med42-v2-70B model~\citep{Christophe2024Med42v2AS}, yielding 400 generated summaries.
Both models use temperature 0, top-$p$ 1, and a maximum completion
length of 2,048 tokens. Our aim is not to rank generation models broadly, but
to determine what evidence-aware evaluation reveals beyond conventional
reference similarity.

\paragraph{Evaluation protocol.}
We run the \textsc{FairSumEval} pipeline with GPT-OSS 120B~\citep{OpenAI2025GPTOSS} and report the complete outcome map together with admission-level Soft IoU-S and Soft IoU-NS. We compare these dimensions with ROUGE-L~\citep{Lin2004ROUGEAP},
sentence-BLEU-4~\citep{Papineni2002BleuAM,Post2018ACF}, BERTScore precision,
recall, and F1~\citep{Zhang2019BERTScoreET}, and
SummaC-ZS~\citep{Laban2021SummaCRN}. We report admission-level means with
95\% bootstrap confidence intervals~\citep{Efron1993Bootstrap} and medians with interquartile ranges,
and quantify associations using Spearman correlation. Undefined Soft IoU
values arising from empty unions are excluded only from the corresponding
analysis.

\vspace{-1mm}
\subsection{Physician-Written and Generated Summaries Differ Sharply in
Evidence Composition}
\label{sec:evidence-composition}

Figure~\ref{fig:motivation} reveals a pronounced asymmetry between the
physician-written reference and the generated summaries. Of the 33,522 facts
extracted from physician-written summaries, only 6,301 (18.8\%) have identified
support in the prior evidence, whereas 27,221 (81.2\%) do not. The generated
summaries exhibit the opposite composition: 6,814 of 10,417 facts (65.4\%) from
the general-purpose model and 7,560 of 11,900 facts (63.5\%) from the medically
specialized model are evidence-supported. Thus, the three sources contain
comparable numbers of supported facts, but the physician references contain
substantially more content without identified prior support (27,221 facts,
compared with 3,603 and 4,340 in the two model outputs).

Importantly, absence of identified support does not imply that a physician
statement is incorrect. Such content may reflect discharge-time decisions,
patient interactions, follow-up planning, or clinical synthesis not documented
in the evidence available to the generation model. Nevertheless, this
asymmetry makes aggregate reference agreement fundamentally ambiguous: failure
to match the reference may indicate either omission of available clinical
information or inability to reproduce information not present in the model
input. These cases are indistinguishable under conventional reference-based
evaluation, motivating \textsc{FairSumEval}'s explicit separation of agreement
with and without evidence support.

\vspace{-1mm}
\subsection{The Outcome Map Reveals Matches, Omissions, and Additions}
\label{sec:outcome-map-results}

Figure~\ref{fig:fairsum_outcome_map} decomposes agreement into matches,
omissions, and additions according to evidence support. Of the 6,301 supported
reference facts, the general purpose model matches 2,091 and omits 4,210,
while the medical model matches 2,700 and omits 3,601. Conversely, 3,240 of
the general purpose model's 6,814 supported facts match the reference, leaving
3,574 additions. For the medical model, 2,731 of 7,560 supported facts match,
leaving 4,829 additions. Both models therefore recover part of the supported
physician content while also selecting supported information absent from the
reference.

The contrast is substantially sharper for facts without identified support in
$E$. Physician references contain 27,221 such facts, more than six times the
3,603 and 4,340 found in the two model outputs. The general purpose and medical
models match only 516 and 650 of these reference facts, leaving 26,705 and
26,571 omissions. In the reverse direction, only 946 and 635 generated facts
match the reference, leaving 2,657 and 3,705 additions. Thus, generations
contain far less content without identified evidence support, and the content
they do contain overlaps only minimally with the physician reference.

The admission level Soft IoUs summarize this separation
(Table~\ref{tab:admission-level-soft-iou-compact}). Soft IoU-S is nearly an
order of magnitude higher than Soft IoU-NS for both the general purpose model
(0.255 vs.\ 0.030) and the medical model (0.236 vs.\ 0.026). Agreement is
therefore concentrated in supported content, whereas content without identified
support is dominated by a large reference only pool and minimal overlap. These
outcomes characterize differences between the reference and generation; they
do not establish that individual omissions or additions are clinically
incorrect.

\vspace{-1mm}
\subsection{Evidence-Aware Evaluation Reveals Two Nonredundant Dimensions}
\label{sec:nonredundant}

Although Soft IoU-S is consistently higher than Soft IoU-NS, this difference
in magnitude does not establish whether the two scores capture the same
underlying behavior or rank admissions similarly. Soft IoU-S summarizes
supported matches, omissions, and additions, while Soft IoU-NS summarizes the
corresponding outcomes for facts without identified support in $E$.
Figure~\ref{fig:soft-iou-relationship} therefore tests whether admissions with
greater evidence supported agreement also exhibit greater agreement on content
without identified support. The association is weak for both the general
purpose model ($\rho=0.22$, $N=200$) and the medical model
($\rho=0.17$, $N=199$), showing that performance in one evidence stratum
provides little information about performance in the other. Combining the two
scores would obscure this variation. \textsc{FairSumEval} therefore reports
Soft IoU-S and Soft IoU-NS separately to preserve these nonredundant views of
agreement between the reference and generated summary.

\vspace{-1mm}
\subsection{
Conventional Metrics Do Not Recover the Evidence-Aware Dimensions}
\label{sec:conventional-metrics}

We next ask whether conventional metrics can recover either evidence aware
dimension without explicitly conditioning on evidence support. As shown in
Figure~\ref{fig:conventional-metric-correlations}, ROUGE-L and BLEU-4 have
substantially higher correlations with Soft IoU-NS than with Soft IoU-S for
both models. For the general purpose model, the correlations are 0.49 versus
0.10 for ROUGE-L and 0.43 versus 0.04 for BLEU-4. The same pattern holds for
the medical model, with correlations of 0.42 versus 0.21 for ROUGE-L and 0.36
versus $-0.06$ for BLEU-4. Lexical similarity therefore aligns more closely
with reproduction of reference content without identified evidence support
than with recovery of evidence supported content. This result is particularly
consequential because facts without identified support constitute most of the
physician reference.

The remaining metrics provide no consistent substitute. BERTScore exhibits the
same pattern for the general purpose model but similar correlations with the
two dimensions for the medical model. SummaC-ZS is weakly correlated with both,
because consistency of a generated summary with the evidence does not determine
whether it selects the same supported facts as the physician reference. Thus,
conventional metrics neither consistently isolate evidence supported agreement
nor reveal when a score is driven by content not identifiable in the model
input. Explicitly reporting Soft IoU-S and Soft IoU-NS exposes this distinction.

\subsection{Validation and Robustness}
We assess the reliability of the three stages of FairSumEval through a clinician study: atomic fact extraction, evidence-support assessment, and cross-summary matching.
This clinician study is complemented by a larger-scale independent blinded LLM
audit. Both show strong agreement with the automated judgments
(Appendices~\ref{app:clinical_evaluation} and~\ref{app:llm-audit}).
Repeating the evaluation across three evaluator seeds, with
generated summaries and evidence held fixed, yields highly stable
results (Appendix~\ref{app:seed_stability}). Extending the LLM analysis
to Gemma-3-27B~\citep{GemmaTeam2025Gemma3} and MedGemma-27B~\citep{Sellergren2025MedGemma} reproduces the evidence-support
asymmetry and the separation between Soft IoU-S and Soft IoU-NS
(Appendix~\ref{app:other_models}). Together, these analyses support
the reliability of the evaluation pipeline and show that the
central findings persist across evaluator seeds and generation
model families.
\vspace{-2mm}
\section{Discussion and Limitations}
\label{sec:discussion}

\paragraph{Aligning evaluation with the intended capability.}
Discharge summaries combine summarization of available evidence
with content involving clinical inference and discharge-time
planning. Evaluation should reflect which capability a method
aims to improve. For evidence-grounded summarization, agreement
on supported content is the relevant focus; for broader discharge-summary
generation, agreement on content without identified support
provides an additional evaluation dimension. Reporting both
separately allows researchers to assess progress against the
intended objective without obscuring performance on the other.
This decomposition provides a clearer basis for developing methods
that address the different demands of discharge-summary generation.

\paragraph{Interpreting disagreement in light of evidence.}
The physician reference is clinically valuable, but it may include
information unavailable to the model and exclude useful facts
present in the evidence. Consequently, omitting a reference fact
does not always indicate a failure to summarize available information,
and adding a fact beyond the reference does not necessarily constitute
hallucination. \textsc{FairSumEval}'s outcome map distinguishes these
cases by locating omissions and additions relative to evidence
support. This context is essential for interpreting overlap:
supported additions may contribute useful information even when
they reduce similarity to the reference.

\paragraph{Limitations and future directions.}
Our study covers 200 MIMIC-III admissions and four models; broader
evaluation would assess generality across institutions, specialties,
and model families. Automated extraction and assessment can introduce
errors and depend on evaluator choice, fact granularity, and evidence
chunking. Support labels are relative to the supplied evidence,
so content without identified support may still reflect valid
clinical synthesis or information documented elsewhere. Future
work should examine sensitivity to these choices and expand the clinician validation (Appendix~\ref{app:clinical_evaluation}) to measure inter-rater agreement and assess the clinical importance of the identified differences.

\vspace{-2mm}
\section{Conclusion}
\label{sec:conclusion}

Improving discharge-summary generation requires evaluating the
right target. \textsc{FairSumEval} introduces an evidence-aware
evaluation framework that distinguishes reference agreement on
facts supported by the model input from agreement on facts without
identified support. Its complete outcome map and two complementary
metrics, Soft IoU-S and Soft IoU-NS, reveal distinctions that
conventional metrics do not consistently capture. This reframing
provides a clearer foundation for developing better generation
methods: progress should be assessed with explicit knowledge of
what information was available and what kind of agreement a score
rewards.

\vspace{-2mm}
\section{Acknowledgements}
\label{sec:ack}
We thank Jung-Oh Lee, Peter Lais, Jack Markert, Meghnil Chowdhury and
Kai Wessendorf for their valuable
clinical insights and thoughtful feedback on this work.
This work was supported by the Institute of Information
and Communications Technology Planning and Evaluation
(IITP) with a grant funded by the Ministry of Science
and ICT (MSIT) of the Republic of Korea in connection
with the Global AI Frontier Lab International Collaborative
Research Agreement (Nos. RS-2024-00469482 and
RS-2024-00509257).

\bibliography{neurips_2026}

\appendix

\clearpage
\raggedbottom
\appendix

\section{FairSumEval Prompts}
\label{app:prompts}

This appendix provides the prompts and structured interfaces used for the
three language model tasks in \textsc{FairSumEval}: atomic fact extraction,
evidence support assessment, and directional cross summary matching. All
tasks require JSON only responses. Evidence support and cross summary
matching require exact supporting quotations, which make positive decisions
auditable. The prompts prohibit outside knowledge and require every meaningful
detail of a fact to be supported.

\lstdefinestyle{fairsumprompt}{
  basicstyle=\ttfamily\scriptsize,
  backgroundcolor=\color{black!2},
  breaklines=true,
  breakatwhitespace=false,
  breakindent=0.8em,
  columns=fullflexible,
  keepspaces=true,
  frame=single,
  framerule=0.35pt,
  framesep=3pt,
  rulecolor=\color{black!25},
  linewidth=\linewidth,
  xleftmargin=0pt,
  xrightmargin=0pt,
  aboveskip=0.5em,
  belowskip=0.8em,
  showstringspaces=false
}

\subsection{Atomic Fact Extraction}
\label{app:fact-extraction-prompt}

Facts are extracted independently from the physician reference and generated
summary. The extractor is instructed to recover every explicitly stated fact,
split compound statements, preserve clinically important qualifiers, and
avoid inference or outside knowledge. We use prompt version
\path{single_pass_domain_general_atomic_v3}.

\par\medskip
\noindent{\small\bfseries Prompt.}\par\nobreak
\begin{lstlisting}[style=fairsumprompt]

You are extracting factual content from a text chunk for evaluation.

Return valid JSON only. Do not use markdown or explanations.

Return exactly this JSON shape:
{
  "facts": [
    {"fact": "one atomic fact"},
    {"fact": "another atomic fact"}
  ]
}

Task:
Extract every explicit atomic fact stated in the text chunk.

Atomic fact definition:
An atomic fact is one self-contained claim about one subject, with one main relation, attribute, event, value, or status.

Rules:
- Extract only facts explicitly stated in the chunk.
- Do not infer, generalize, explain, or add outside knowledge.
- Do not summarize a paragraph as one broad fact.
- Do not merge multiple facts into one item.
- If a sentence contains a list of separate attributes, values, findings, items, dates, people, places, actions, or outcomes, split them into separate facts.
- Keep necessary context in each fact so it can stand alone.
- Preserve numbers, units, dates, names, negation, and qualifiers when they matter.
- Do not create vague facts such as "the document discusses several issues."
- If the chunk contains no explicit factual claims, return {"facts": []}.

Do not copy task instructions or examples into the output.
\end{lstlisting}

The following task metadata and summary chunk are appended at call time.

\par\medskip
\noindent{\small\bfseries Input format.}\par\nobreak
\begin{lstlisting}[style=fairsumprompt]
Return exactly this JSON shape:
{"facts":[{"fact":"one atomic fact"}, {"fact":"another atomic fact"}]}

prompt_version: single_pass_domain_general_atomic_v3
source_kind: <generated|ground_truth>

SUMMARY CHUNK:
<chunk text>
\end{lstlisting}

\par\medskip
\noindent{\small\bfseries Output format.}\par\nobreak
\begin{lstlisting}[style=fairsumprompt]
{
  "facts": [
    {"fact": "one atomic fact"},
    {"fact": "another atomic fact"}
  ]
}
\end{lstlisting}

\subsection{Evidence Support Assessment}
\label{app:evidence-support-prompt}

Each extracted fact is assessed against the clinical evidence available to
the generation model. Support requires every meaningful detail to be stated
or unambiguously paraphrased in the supplied evidence. Clinical plausibility,
external knowledge, and speculative inference do not qualify as support. When
the evidence is divided into chunks, a fact is labeled supported if at least
one chunk fully supports it.

\par\medskip
\noindent{\small\bfseries Prompt.}\par\nobreak

\begin{lstlisting}[style=fairsumprompt]
You are a strict evidence-support judge.
Return JSON only. Use only the provided evidence text for each task. Do not use outside knowledge.

Classify how the evidence supports the exact factual content of each claim.

This is not a semantic-similarity task. Pay close attention to negation, status,
quantities, values, units, dates, times, entities, and qualifiers.

Labels:
- supported: every meaningful detail is explicitly stated or unambiguously paraphrased in the supplied evidence.
- not_supported: at least one meaningful detail is absent, contradicted, changed, overly specific, or requires speculative inference.

For supported, include evidence_quote as an exact
substring copied from the evidence that carries the support.
For not_supported, use an empty evidence_quote.
Return one prediction per claim and nothing else.
\end{lstlisting}

\par\medskip
\noindent{\small\bfseries Input format.}\par\nobreak
\begin{lstlisting}[style=fairsumprompt]
{
  "verification_kind": "support",
  "evidence": "<evidence text>",
  "claims": [
    {"claim_id": "<claim identifier>", "claim": "<atomic fact>"}
  ]
}
\end{lstlisting}

\par\medskip
\noindent{\small\bfseries Output format.}\par\nobreak
\begin{lstlisting}[style=fairsumprompt]
{
  "predictions": [
    {
      "claim_id": "<claim identifier>",
      "label": "<supported|not_supported>",
      "evidence_quote": "<exact evidence substring or empty string>"
    }
  ]
}
\end{lstlisting}

\subsection{Cross-Summary Matching}
\label{app:cross-summary-prompt}

Cross summary matching uses the same strict factual support criterion, with
the opposite summary supplied as evidence. Reference facts are checked against
the complete generated summary, and generated facts are checked against the
complete reference. Matching is therefore directional and does not impose a
one to one correspondence between independently extracted facts.

\par\medskip
\noindent{\small\bfseries Prompt.}\par\nobreak
\begin{lstlisting}[style=fairsumprompt]
You are a strict factual support judge.
Return JSON only. Use only the provided evidence text for each task. Do not use outside knowledge.

Decide whether the evidence fully supports the exact factual content of the claim.

This is not a semantic-similarity task. The evidence must support every meaningful part of the claim.
Pay close attention to negation, status, quantities, values, units, dates, times, entities, and qualifiers.

Labels:
- fully_supported: every meaningful part of the claim is explicitly supported by the evidence.
- not_fully_supported: in the evidence are meaningful parts missing, changed, weakened, contradicted.

If the evidence is topically related but incomplete, the label is not_fully_supported.
For fully_supported claims, include evidence_quote as a exact substring copied from the evidence. Pay attention that the quote provides
enough information to support/entail the claim.
For not_fully_supported claims, use an empty evidence_quote.
Return one prediction per claim and nothing else.
\end{lstlisting}

\par\medskip
\noindent{\small\bfseries Input format.}\par\nobreak
\begin{lstlisting}[style=fairsumprompt]
{
  "verification_kind": "cross_summary",
  "direction": "<reference_to_generation|generation_to_reference>",
  "evidence": "<complete opposite summary>",
  "claims": [
    {"claim_id": "<claim identifier>", "claim": "<atomic fact>"}
  ]
}
\end{lstlisting}

\noindent\begin{minipage}{\columnwidth}
\par\medskip
\noindent{\small\bfseries Output format.}\par\nobreak
\begin{lstlisting}[style=fairsumprompt]
{
  "predictions": [
    {
      "claim_id": "<claim identifier>",
      "label": "<fully_supported|not_fully_supported>",
      "evidence_quote": "<exact summary substring or empty string>"
    }
  ]
}
\end{lstlisting}
\end{minipage}

\subsection{Implementation Details}
\label{app:implementation-details}

We run all three tasks using GPT-OSS~\citep{OpenAI2025GPTOSS}. Facts are
extracted independently from each chunk of the physician reference and
generated summary. Each fact is then assessed for support in the prior
clinical evidence. When the evidence is chunked, the fact is considered
supported if any chunk fully supports it. Cross summary matching is performed
in both directions using the complete opposite summary. Responses are parsed using the task specific JSON schemas above.

The full evaluator configuration is as follows:
\begin{itemize}
    \item Evaluator checkpoint: \texttt{openai/gpt-oss-120b} (Hugging Face revision \texttt{b5c939de8f754692c1647ca79fbf85e8c1e70f8a}; served alias \texttt{openai/gpt-oss-120b}).
    \item Decoding parameters: temperature $=0.0$, top-$p$ $=1.0$, seed $=123$, reasoning effort $=\texttt{low}$, maximum output length $=12{,}000$ tokens; request timeout $=1{,}800$\,s; support/cross batch size $=20$.
    \item Summary chunking: fact extraction over boundary-aware chunks of at most $4{,}000$ characters with \emph{no overlap} (contiguous splits at paragraph/sentence/whitespace boundaries).
    \item Evidence chunking: whole-document packing up to $96{,}000$ characters per evidence unit (profile \texttt{direct\_joint\_semantic\_only\_whole\_document}); no overlap. A fact is supported if any evidence unit fully supports it.
    \item Structured-output handling: responses must parse as the task JSON schema; up to $3$ JSON parse attempts per call (invalid JSON triggers a ``return only one JSON object'' retry with a bumped seed offset); HTTP transport retries up to $4$ times with exponential backoff ($5$\,s base) on transient failures; permanent client errors (HTTP $400/401/403/404$) and length-truncated outputs are not retried as successes. Unresolved / invalid judgments are recorded as failures and do not default to supported or matched.
\end{itemize}

\paragraph{Outcome construction.}
For each fact, \textsc{FairSumEval} combines its source, evidence support label,
and directional matching label. A reference fact unmatched by the generated
summary is an omission, whereas a generated fact unmatched by the reference is
an addition. These outcomes are separated according to whether the assessed
fact has identified support in the evidence. Directional matches, omissions,
and additions within the supported stratum determine Soft IoU-S; the
corresponding outcomes among facts without identified support determine Soft
IoU-NS. A zero denominator makes the affected score undefined, and that
admission is excluded only from analyses requiring the undefined score.
All facts receive equal weight; clinical-importance weights could instead
replace unit counts in the outcome map and Soft IoU metrics in future work.

\section{Supplementary Preference Analysis}
\label{app:preference-analysis}

Before consolidating the main analysis around the complete outcome map, we
conducted an exploratory blinded language model preference study on the same
200 admissions. For each admission, the evaluator compared the two generated
summaries under five objectives: supported omissions, omissions without
identified evidence support, supported additions, additions without identified
evidence support, and overall clinical quality. This analysis is included as
supplementary evidence that model preference depends on the evaluation
objective; it is not used as clinical validation of \textsc{FairSumEval}.

\begin{table}[h!]
    \centering
    \small
    \caption{\textbf{Objective specific model preferences.} Counts show the
    number of admissions for which the blinded language model evaluator
    preferred each generation model or returned a tie. The preferred model
    changes across objectives, while the overall quality comparison produces
    no clear winner.}
    \label{tab:appendix-preferences}
    \begin{tabular}{lrrr}
        \toprule
        Evaluation objective & Llama & Med42 & Tie \\
        \midrule
        Fewer supported omissions & 48 & 98 & 54 \\
        Fewer no support omissions & 41 & 53 & 106 \\
        Fewer supported additions & 159 & 21 & 20 \\
        Fewer no support additions & 135 & 61 & 4 \\
        Overall clinical quality & 77 & 86 & 37 \\
        \bottomrule
    \end{tabular}
\end{table}

Preferences also changed within individual admissions. Among the 135
admissions with decisive judgments for both supported omissions and supported
additions, the preferred model reversed in 102 cases (75.6\%). Overall clinical
quality yielded no clear winner: Med42 was preferred in 86 admissions, Llama
in 77, and 37 were ties ($p=0.53$, two sided exact binomial test among decisive
judgments). These results complement the weak association between Soft IoU-S
and Soft IoU-NS in the main paper: conclusions about model quality depend on
which evidence aware behavior is prioritized.

\section{Clinician Validation of \textsc{FairSumEval} Stages}
\label{app:clinical_evaluation}

Five clinicians validated the three stages of \textsc{FairSumEval}: atomic
fact extraction, evidence-support assessment, and cross-summary matching.
Each clinician independently reviewed a different admission. Evaluated items
were balanced across \textsc{FairSumEval} labels (support in $E$ vs.\ no
support in $E$; match vs.\ non-match) rather than sampled at their natural
prevalence. Clinicians could mark any decision as uncertain.

\paragraph{Atomic fact extraction.}
Each clinician reviewed one complete discharge summary together with all
facts extracted from it and rated faithfulness and completeness on a 1--5
scale, where 5 indicates that the facts faithfully represent all clinically
meaningful content with no meaningful fact missing.

\paragraph{Evidence support.}
Each clinician judged four facts against the corresponding prior clinical
evidence, labeling a fact as supported only if every meaningful detail was
directly supported. Facts were drawn from both physician references and
generated summaries.

\paragraph{Cross-summary matching.}
Each clinician judged four facts against the opposite summary, deciding
whether all meaningful information in the fact was present. Facts covered
both matching directions.

\paragraph{Results.}
Table~\ref{tab:clinician-validation} summarizes the results. Clinicians rated
extraction highly (mean 4.8; median 5; range 4--5), with all five summaries
rated at least 4. Agreement with \textsc{FairSumEval} was 19/20 (95\%) for
evidence support, including 10/10 support and 9/10 no-support decisions, and
19/20 (95\%) for cross-summary matching, including 10/10 match and 9/10
non-match decisions. No judgments were marked uncertain. These results are
consistent with reliable stage-level decisions throughout the pipeline that
constructs the outcome map and Soft IoU scores, and they complement the
larger blinded audit in Appendix~\ref{app:llm-audit}. The Wilson intervals
are wide, reflecting the small sample, and because sampling was balanced,
agreement describes this stratified sample rather than uncertainty in the
cohort-level 18.8\% reference-support estimate. Because clinicians reviewed
distinct admissions, inter-rater agreement could not be estimated; a planned
retrospective hospital study with overlapping clinician assessments will
measure it.

\begin{table}[!t]
\centering
\scriptsize
\caption{\textbf{Clinician validation of \textsc{FairSumEval} stage
decisions.} Five clinicians evaluate each stage on distinct admissions,
with items balanced across \textsc{FairSumEval} labels. Agreement is
computed over non-uncertain responses; binary estimates report 95\%
Wilson confidence intervals~\citep{Wilson1927ProbableInference}.}
\label{tab:clinician-validation}
\setlength{\tabcolsep}{2pt}
\renewcommand{\arraystretch}{1.15}
\begin{tabular}{@{}llr@{}}
\toprule
\textbf{Stage} & \textbf{Quantity} & \textbf{Result} \\
\midrule

Extraction
    & Mean rating & 4.8 \\
    & Median [range] & 5 [4--5] \\
    & Ratings $\geq 4$ & 5/5 \\

\midrule

Evidence support
    & Overall agreement & 19/20 (95.0\% [76.4, 99.1]) \\
    & Support in $E$ & 10/10 (100\% [72.2, 100]) \\
    & No support in $E$ & 9/10 (90.0\% [59.6, 98.2]) \\
    & Uncertain & 0 \\

\midrule

Matching
    & Overall agreement & 19/20 (95.0\% [76.4, 99.1]) \\
    & Match & 10/10 (100\% [72.2, 100]) \\
    & Non-match & 9/10 (90.0\% [59.6, 98.2]) \\
    & Uncertain & 0 \\

\bottomrule
\end{tabular}
\end{table}

\section{Blinded LLM Audit of \textsc{FairSumEval} Stages}
\label{app:llm-audit}


To validate the same three stages at larger scale, we use Claude Opus 4.6~\citep{Anthropic2026Opus46}
as an independent, blinded evaluator on the 200-admission cohort. The
evaluator is not shown the \textsc{FairSumEval} label, the generation model
identity, or the downstream outcome assigned to the fact.
Table~\ref{tab:llm-audit} reports estimates with sample sizes.

\paragraph{Atomic fact extraction.}
We sample 20 summaries (10 physician references, 5 Llama-3.3-70B, and
5 Med42-v2-70B). The evaluator reads each summary and its extracted fact
list and rates faithfulness from 1 to 5. Ratings are high
(mean $4.3\pm0.7$; median 4; range 2--5), with 19/20 summaries rated
$\geq 4$.

\paragraph{Evidence support.}
The evaluator receives only the atomic fact and the corresponding prior
clinical evidence and decides whether every meaningful detail is supported.
We stratify by \textsc{FairSumEval} label ($n{=}797$). Agreement is 89.9\%
on facts labeled support in $E$ and 86.0\% on facts labeled no support in
$E$ (overall 88.0\%).

\paragraph{Cross-summary matching.}
The evaluator receives only the fact and the opposite summary and decides
whether the fact is stated in that summary ($n{=}797$). Agreement is 87.8\%
on match and 90.1\% on non-match decisions (overall 89.5\%).

\paragraph{Scope.}
These agreements support the reliability of the stage-wise labels used to
construct Soft IoU-S and Soft IoU-NS. They do not establish clinical
correctness: a fact without identified support may still be clinically
reasonable, inferred, or documented outside the evidence supplied to the
model. Agreement between models complements, but does not replace, the
clinician validation in Appendix~\ref{app:clinical_evaluation}.

\begin{table}[h!]
    \centering
    \scriptsize
    \setlength{\tabcolsep}{2pt}
    \caption{\textbf{Blinded LLM audit of \textsc{FairSumEval} stage
    decisions.} Independent Claude Opus 4.6 judgments versus
    \textsc{FairSumEval} labels. Rating mean$\pm$SD is over summaries;
    binary estimates report 95\% Wilson confidence intervals~\citep{Wilson1927ProbableInference}.}
    \label{tab:llm-audit}
    \begin{tabular}{llrc}
        \toprule
        Stage & Quantity & Estimate & $n$ \\
        \midrule
        Extraction
          & Rating mean $\pm$ SD
          & $4.3 \pm 0.7$
          & 20 \\
          & Median [range]
          & $4$ $[2, 5]$
          & 20 \\
          & Ratings $\geq 4$
          & $19/20$
          & 20 \\
        \midrule
        Evidence support
          & Overall agreement
          & 88.0\% [85.5, 90.0]
          & 797 \\
          & Support in $E$
          & 89.9\% [86.6, 92.5]
          & 397 \\
          & No support in $E$
          & 86.0\% [82.3, 89.1]
          & 400 \\
        \midrule
        Matching
          & Overall agreement
          & 89.5\% [87.1, 91.4]
          & 797 \\
          & Match
          & 87.8\% [82.8, 91.5]
          & 221 \\
          & Non-match
          & 90.1\% [87.4, 92.3]
          & 576 \\
        \bottomrule
    \end{tabular}
\end{table}

\section{Generalization to Gemma and MedGemma}
\label{app:other_models}
\begin{figure*}[!t]
     \vspace{-2mm}
    \centering
    \includegraphics[
        width=\textwidth,
        trim=0 0 0 0,
        clip
    ]{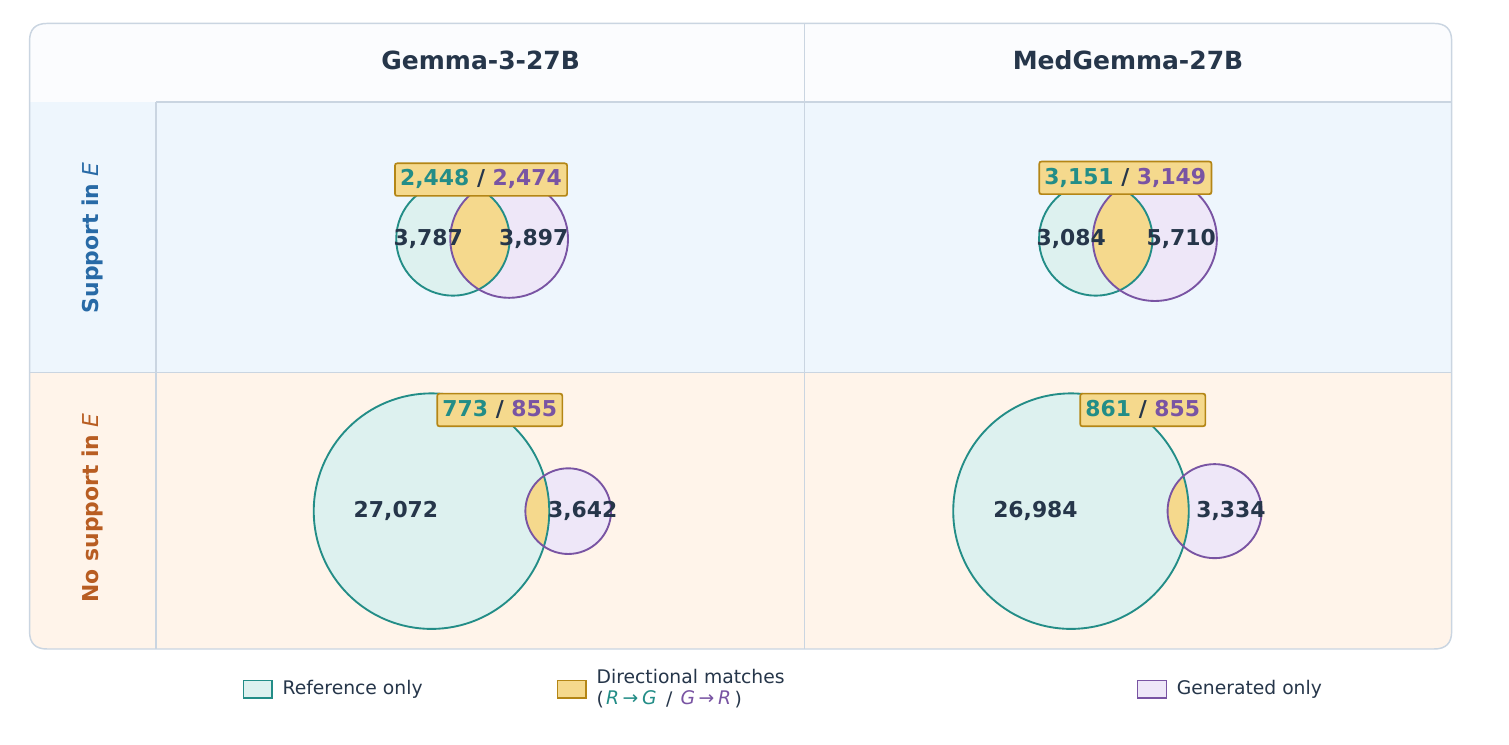}
    \vspace{-5mm}
\caption{\textbf{The evidence-aware outcome pattern generalizes to Gemma and MedGemma.}
Rows separate facts with identified support in $E$ from those without
identified support. Green, gold, and purple regions denote
reference-only facts, directional matches, and generation-only facts,
respectively. As in the main-paper experiments, both models show
substantially greater overlap within the supported stratum, whereas
content without identified support is dominated by a large
reference-only pool. Gold labels report directional match counts as
$R\!\rightarrow\!G / G\!\rightarrow\!R$. All counts are pooled across
200 admissions.}
    \label{fig:outcome-map-gemma}
\end{figure*}

\begin{table*}[!t]
     \vspace{-3mm}
    \footnotesize
    \centering
\caption{\textbf{Admission-level Soft IoU on Gemma and MedGemma confirms the same evidence-aware separation.}
Repeating the main-paper experiment with an additional
general-purpose/medical model pair yields the same separation between
the two evidence-aware dimensions. Soft IoU-S is substantially higher
than Soft IoU-NS for both Gemma (0.239 vs.\ 0.032) and MedGemma
(0.249 vs.\ 0.033), showing that the main-paper finding generalizes
beyond Llama and Med42. Columns report means [95\% bootstrap CI] and
medians $[Q_1,Q_3]$.}
\vspace{-2mm}
    \label{tab:soft-iou-gemma-medgemma}
    \setlength{\tabcolsep}{5pt}
    \renewcommand{\arraystretch}{1.18}
    \begin{tabular}{lcccc}
        \toprule
        & \multicolumn{2}{c}{$\mathbf{Soft\ IoU\text{-}S}\;\boldsymbol{\uparrow}$}
            & \multicolumn{2}{c}{$\mathbf{Soft\ IoU\text{-}NS}\;\boldsymbol{\uparrow}$} \\
        \cmidrule(lr){2-3}\cmidrule(lr){4-5}
        \textbf{Generation model}
            & Mean
            & Median
            & Mean
            & Median \\
        \midrule
        Gemma-3-27B
            & 0.239 {[0.224, 0.254]}
            & 0.235 {[0.173, 0.295]}
            & 0.032 {[0.028, 0.036]}
            & 0.026 {[0.015, 0.039]} \\
        MedGemma-27B
            & \textbf{0.249 {[0.230, 0.269]}}
            & \textbf{0.239 {[0.146, 0.335]}}
            & \textbf{0.033 {[0.029, 0.037]}}
            & \textbf{0.026 {[0.014, 0.045]}} \\
        \bottomrule
    \end{tabular}
\end{table*}

\begin{figure*}[!t]
     \vspace{-2mm}
    \centering
    \includegraphics[
        width=0.7\textwidth,
        trim=0 0 0 0,
        clip
    ]{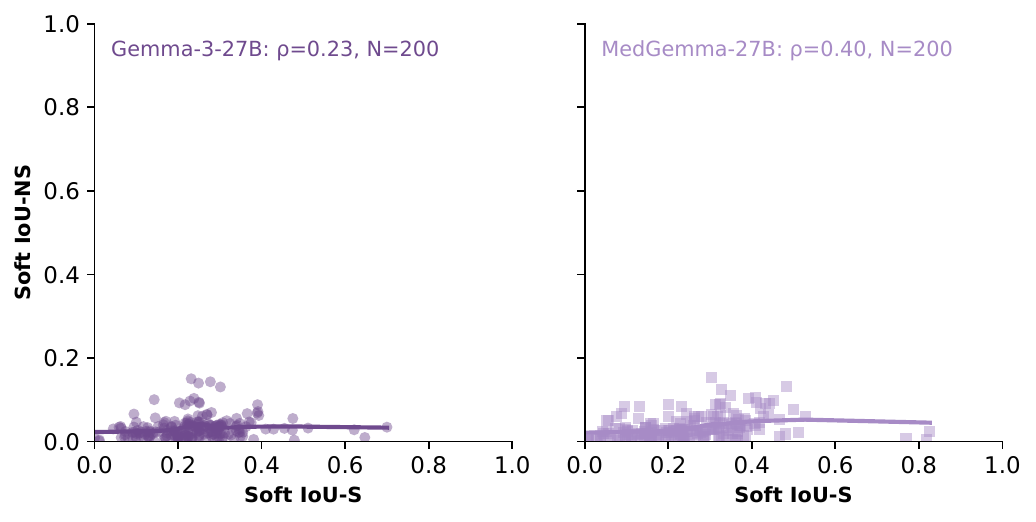}
    \vspace{-5mm}
\caption{\textbf{The non-redundancy of Soft IoU-S and Soft IoU-NS
generalizes to additional models.}
Each point represents one admission for Gemma-3-27B (left) or
MedGemma-27B (right). Soft IoU-S and Soft IoU-NS measure
reference--generation overlap among facts with and without identified
evidence support, respectively. The two scores remain only partially
associated across admissions (Spearman $\rho=0.23$ and $0.40$),
showing that stronger agreement on supported content does not determine
agreement on content without identified support. This reproduces the
central finding from the main-paper experiments that the two
evidence-aware dimensions should be reported separately. Trend lines
are visual aids only.}
    \label{fig:soft-iou-relationship-gemma}
\end{figure*}

To test whether our findings generalize beyond Llama-3.3-70B and
Med42-v2-70B, we repeat the full \textsc{FairSumEval} analysis on the same
200 admissions with a second general-purpose/medical model pair:
Gemma-3-27B~\citep{GemmaTeam2025Gemma3} and MedGemma-27B~\citep{Sellergren2025MedGemma}.

\paragraph{Evidence composition.}
Generated summaries again consist mostly of evidence-supported content:
6,371 of 10,868 Gemma facts (58.6\%) and 8,859 of 13,048 MedGemma facts
(67.9\%), compared with 6,235 of 34,080 physician-reference facts (18.3\%).
Atomic fact extraction was rerun for the Gemma/MedGemma experiments,
resulting in slightly different reference fact counts from the main
analysis despite using the same 200 admissions.

\paragraph{Outcome map and Soft IoU.}
Both models show substantially greater overlap within the supported stratum,
whereas content without identified support is dominated by a large
reference-only pool (Figure~\ref{fig:outcome-map-gemma}). Soft IoU-S remains
substantially higher than Soft IoU-NS for both Gemma (0.239 vs.\ 0.032) and
MedGemma (0.249 vs.\ 0.033; Table~\ref{tab:soft-iou-gemma-medgemma}). The two
dimensions remain only partially associated across admissions (Spearman
$\rho=0.23$ and $0.40$; Figure~\ref{fig:soft-iou-relationship-gemma}).

Together, these results reproduce the evidence-composition asymmetry, the
gap between supported and no-support overlap, and the non-redundancy of the
two dimensions across four models, two model families, and both
general-purpose and medically specialized generators.

\begin{figure*}[!t]
    \centering
    \includegraphics[width=\textwidth]
        {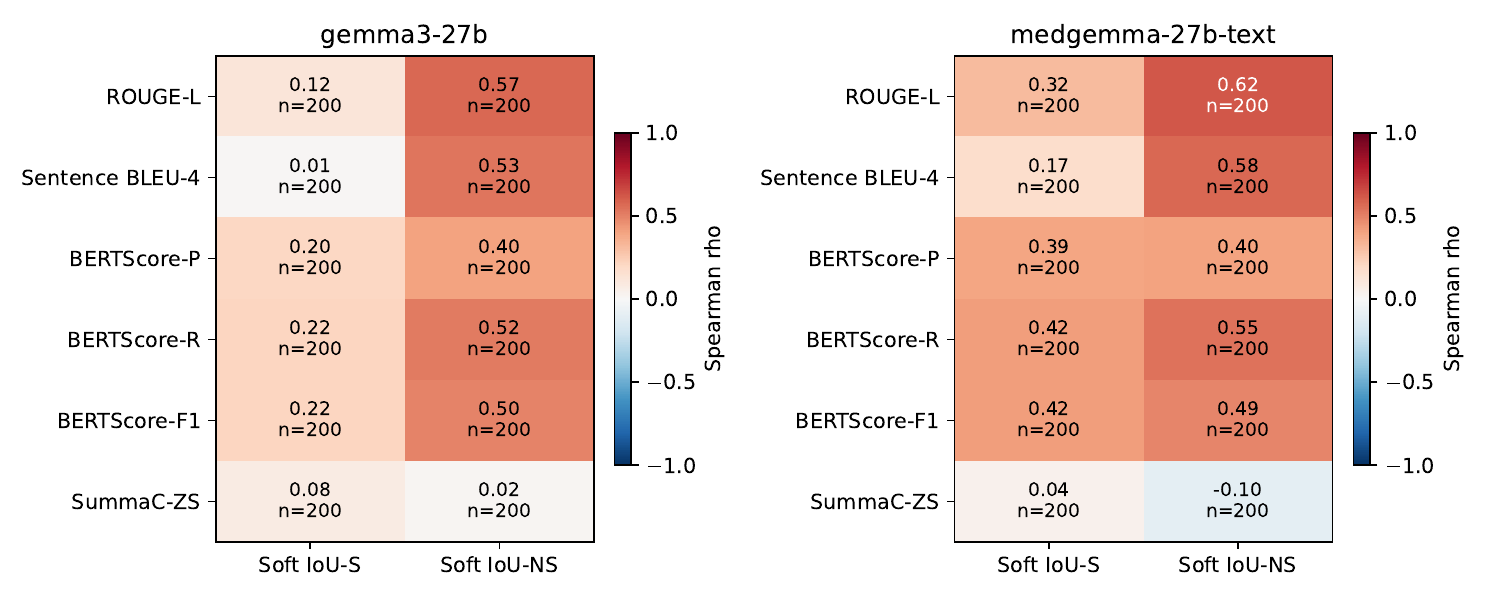}
    \caption{\textbf{
    Conventional metrics do not consistently capture evidence-aware agreement for Gemma and MedGemma.}
    Each cell reports the admission-level Spearman correlation of
    a conventional metric with Soft IoU-S or Soft IoU-NS for
    Gemma-3-27B (left) and MedGemma-27B (right;
    $N=200$ for every cell).
    Columns distinguish reference--generation agreement among
    facts with identified evidence support from agreement among
    facts without identified support.
    ROUGE-L and BLEU-4 correlate more strongly with Soft IoU-NS
    for both models. BERTScore precision, recall and F1 show
    larger differences between the two correlations for Gemma
    than for MedGemma, while SummaC-ZS has weak correlations
    with both dimensions.}
    \label{fig:conventional-metrics-gemma}
\end{figure*}

\paragraph{Relationship to conventional metrics.}
The conventional-metric pattern also extends to Gemma and MedGemma
(Figure~\ref{fig:conventional-metrics-gemma}).
For both models, ROUGE-L and BLEU-4 correlate more strongly with
Soft IoU-NS ($\rho=0.53$--$0.62$) than with Soft IoU-S
($\rho=0.01$--$0.32$).
All three BERTScore components show higher correlations with
Soft IoU-NS for Gemma, whereas the differences are smaller for
MedGemma; its BERTScore precision is similarly correlated with
both dimensions ($0.39$ vs.\ $0.40$).
SummaC-ZS is weakly associated with both dimensions.
These results extend the main-paper findings (Figure 5) to two additional generation models: ROUGE-L and BLEU-4 consistently correlate more strongly with Soft IoU-NS, while BERTScore shows a less consistent pattern and SummaC-ZS remains weakly correlated with both dimensions.

\section{Evaluator Stability Across Seeds}
\label{app:seed_stability}

To assess sensitivity to evaluator stochasticity, we repeat the complete
\textsc{FairSumEval} evaluation with evaluator seeds 123, 456, and 789,
holding the generated summaries and clinical evidence fixed. This analysis
therefore isolates evaluator variability; it does not measure variability
from independently regenerated summaries.

\paragraph{Estimates.}
Evidence-support rates vary by only 0.2--0.5 percentage points across seeds
(Table~\ref{tab:seed-stability-support}), and supported-stratum outcome
counts show similarly small variation
(Table~\ref{tab:seed-stability-outcome-map}). At the admission level, Soft
IoU-S varies by approximately 0.001 for both models and Soft IoU-NS by at
most 0.001 (Table~\ref{tab:seed-stability-soft-iou}).

\paragraph{Relationship between dimensions.}
The ratio of Soft IoU-S to Soft IoU-NS remains nearly unchanged across seeds,
and the admission-level association between the two scores remains weak:
Spearman $\rho=0.19\pm0.03$ for Llama-3.3-70B and $\rho=0.13\pm0.04$ for
Med42-v2-70B (Table~\ref{tab:seed-stability-soft-iou-relation}). The central
\textsc{FairSumEval} findings are therefore insensitive to evaluator seed
under fixed generations and evidence.

\begin{table}[!t]
\centering
\footnotesize
\caption{\textbf{Seed stability of evidence-support rates
(mean $\pm$ SD across evaluator seeds).}
For each evaluator seed, we compute the fraction of facts classified as
supported by evidence $E$. Values report mean $\pm$ SD across three
evaluator seeds (123/456/789), with generated summaries and evidence
held fixed ($n=200$ admissions).}
\label{tab:seed-stability-support}
\setlength{\tabcolsep}{6pt}
\renewcommand{\arraystretch}{1.18}
\begin{tabular}{lc}
\toprule
\textbf{Source} & \textbf{Supported by $E$ (\%)} \\
\midrule
Reference $R$ & $18.5 \pm 0.2$ \\
Llama-3.3-70B ($G$) & $65.1 \pm 0.3$ \\
Med42-v2-70B ($G$) & $64.0 \pm 0.5$ \\
\bottomrule
\end{tabular}
\end{table}

\begin{table*}[!t]
\centering
\footnotesize
\caption{\textbf{Seed stability of supported-stratum outcome-map counts
(mean $\pm$ SD across evaluator seeds).}
For each evaluator seed, we pool supported-stratum directional matches,
reference-only omissions, and generation-only additions across the
200 admissions. Values report mean $\pm$ SD across three evaluator
seeds (123/456/789), with generated summaries and evidence held fixed.}
\label{tab:seed-stability-outcome-map}
\setlength{\tabcolsep}{5pt}
\renewcommand{\arraystretch}{1.18}
\begin{tabular}{lccc}
\toprule
\textbf{Model}
 & \textbf{Matches ($R\!\rightarrow\!G$)}
 & \textbf{Omissions ($R$-only)}
 & \textbf{Additions ($G$-only)} \\
\midrule
Llama-3.3-70B & $2067 \pm 22$ & $4140 \pm 62$ & $3492 \pm 71$ \\
Med42-v2-70B & $2634 \pm 57$ & $3573 \pm 32$ & $4735 \pm 98$ \\
\bottomrule
\end{tabular}
\end{table*}

\begin{table}[!t]
\centering
\footnotesize
\caption{\textbf{Seed stability of Soft IoU scores
(mean $\pm$ SD across evaluator seeds).}
For each evaluator seed, we compute the mean admission-level Soft IoU-S
and Soft IoU-NS across the 200 admissions. The table reports the mean
and SD of these values across three evaluator seeds (123/456/789), with
generated summaries and evidence held fixed.}
\label{tab:seed-stability-soft-iou}
\setlength{\tabcolsep}{6pt}
\renewcommand{\arraystretch}{1.18}
\begin{tabular}{lcc}
\toprule
\textbf{Model} & \textbf{Soft IoU-S} & \textbf{Soft IoU-NS} \\
\midrule
Llama-3.3-70B & $0.257 \pm 0.001$ & $0.030 \pm 0.001$ \\
Med42-v2-70B & $0.236 \pm 0.001$ & $0.025 \pm 0.0003$ \\
\bottomrule
\end{tabular}
\end{table}

\begin{table}[!t]
\centering
\footnotesize
\caption{\textbf{Seed stability of the relationship between Soft IoU-S
and Soft IoU-NS (mean $\pm$ SD across evaluator seeds).}
For each evaluator seed, we compute the ratio of mean Soft IoU-S to mean
Soft IoU-NS and the admission-level Spearman correlation between the two
scores. Values report mean $\pm$ SD across seeds (123/456/789), with
generated summaries and evidence held fixed.}
\label{tab:seed-stability-soft-iou-relation}
\setlength{\tabcolsep}{3pt}
\renewcommand{\arraystretch}{1.18}
\begin{tabular}{@{}lcc@{}}
\toprule
\textbf{Model}
 & \textbf{IoU-S / IoU-NS}
 & \textbf{Spearman $\rho$} \\
\midrule
Llama-3.3-70B & $8.6 \pm 0.1$ & $0.19 \pm 0.03$ \\
Med42-v2-70B & $9.3 \pm 0.1$ & $0.13 \pm 0.04$ \\
\bottomrule
\end{tabular}
\end{table}

\section{Cohort Selection and Evidence Construction}
\label{app:cohort-evidence}

\paragraph{Cohort selection.}
Starting from 58,976 MIMIC-III admissions, we retain 47,006 with exactly one
discharge summary, restrict to the held-out test split (4,809), and keep
admissions with at least one eligible clinical note (4,786). From a
1,000-admission generation subset, we randomly sample 200 admissions
(seed 123) with complete outputs from both generation models and fix this
cohort for all analyses. Table~\ref{tab:cohort} summarizes the selection
flow and cohort characteristics. The admission IDs are available in the code repository.

\paragraph{Evidence construction.}
$E$ comprises the admission's notes from nine MIMIC-III categories
(Physician, Nursing, Nursing/other, Radiology, Echo, ECG, Respiratory,
Nutrition, and General), ordered chronologically with near-duplicates
removed. In consultation with clinicians, we exclude five categories rarely
used when writing discharge summaries (Case Management, Consult, Pharmacy,
Rehab Services, and Social Work); these comprise 0.44\% of MIMIC-III notes
and occur in 12 of the 200 admissions. The physician-written discharge
summary is identified by its unique note ID and held out as the reference
$R$, so it never enters $E$. Discharge-day notes are otherwise retained; we
do not filter by timestamp. Comparing note timestamps with
\texttt{ADMISSIONS.DISCHTIME}, only 5 of 2,265 notes, spanning 4 admissions,
were recorded after discharge; excluding them changes reference support
from 18.80\% to 18.71\%. $E$ contains clinical notes only, not structured
laboratory results or medication orders.

\paragraph{Context budget and truncation.}
Both models receive identical evidence under a common 8,192-token context
(Med42-v2's native limit). After instructions and the 2,048-token output
allowance, approximately 5,650 tokens remain for evidence. The complete
evidence fits for 114 admissions; for the remaining 86, we retain the
earliest notes within the budget, so late-stay and discharge-day notes may
be excluded. \textsc{FairSumEval} assesses support against exactly the
evidence supplied to the models, so support labels are consistent with the
model input by construction. All 400 generated summaries complete within
the output budget.

\paragraph{Robustness to truncation.}
Restricting to the 114 admissions in which the models receive the complete
evidence, including all discharge-day notes, only 14.3\% of
physician-reference facts have identified support, compared with 67.3\%
and 62.3\% of generated facts for Llama and Med42. Soft IoU-S remains
substantially higher than Soft IoU-NS for both models (0.260 vs.\ 0.026 for
Llama; 0.233 vs.\ 0.025 for Med42). The evidence-support asymmetry and
separation between Soft IoU-S and Soft IoU-NS therefore persist even when
all available clinical notes fit within the model context. Incorporating medication orders and other
structured EHR data is a natural next step, and \textsc{FairSumEval}
directly quantifies how much additional reference content each new
evidence source supports.

\begin{table}[!t]
\centering
\scriptsize
\caption{\textbf{Cohort selection and evidence characteristics.}
Selection flow from MIMIC-III to the 200-admission evaluation cohort,
cohort demographics, and the size of the evidence $E$ per admission.}
\label{tab:cohort}
\setlength{\tabcolsep}{3pt}
\renewcommand{\arraystretch}{1.12}
\begin{tabular}{@{}lr@{}}
\toprule
\textbf{Selection step} & \textbf{Admissions} \\
\midrule
MIMIC-III admissions & 58,976 \\
Exactly one discharge summary & 47,006 \\
Held-out test split & 4,809 \\
$\geq 1$ eligible clinical note & 4,786 \\
Generation subset & 1,000 \\
Evaluation cohort & 200 \\
\midrule
\textbf{Cohort characteristic} & \textbf{Value} \\
\midrule
Patients & 194 (6 with two admissions) \\
Median age (years) & 63.1 \\
Male & 60.5\% \\
Emergency admissions & 79.5\% \\
Newborn admissions & 14 \\
\midrule
\textbf{Evidence $E$ per admission} & \textbf{Value} \\
\midrule
Notes, median [IQR] & 13 [8--23] \\
Admissions with $\leq 3$ notes & 18 \\
Tokens, median [IQR] & 4,510 [2,172--11,716] \\
Tokens, maximum & 133,640 \\
Complete $E$ fits in context & 114 \\
\bottomrule
\end{tabular}
\end{table}

\end{document}